\def\year{2018}\relax
\documentclass[letterpaper]{article} 
\usepackage{aaai18}  
\usepackage{times}  
\usepackage{helvet}  
\usepackage{courier}  
\usepackage{url}  
\usepackage{graphicx}  
\usepackage{algorithm}
\usepackage{algorithmic}
\usepackage{amsmath}
\usepackage{bbold}
\usepackage{subfigure}
\usepackage{bm}

\begin{document}
%
\title{Variational Probability Flow for \\ Biologically Plausible Training of Deep Neural Networks}
\author{Zuozhu Liu \and Tony Q.S. Quek \and Shaowei Lin \\
Singapore University of Technology and Design\\
8 Somapah Road, Singapore, 487372\\
zuozhu\_liu@mymail.sutd.edu.sg,  \{tonyquek, shaowei\_lin\}@sutd.edu.sg
}

\maketitle
\begin{abstract}
The quest for biologically plausible deep learning is driven, not just by the desire to explain experimentally-observed properties of biological neural networks, but also by the hope of discovering more efficient methods for training artificial networks. In this paper, we propose a new algorithm named Variational Probably Flow (VPF), an extension of minimum probability flow for training binary Deep Boltzmann Machines (DBMs). We show that weight updates in VPF are local, depending only on the states and firing rates of the adjacent neurons. Unlike contrastive divergence, there is no need for Gibbs confabulations; and unlike backpropagation, alternating feedforward and feedback phases are not required. Moreover, the learning algorithm is effective for training DBMs with intra-layer connections between the hidden nodes. Experiments with MNIST and Fashion MNIST demonstrate that VPF learns reasonable features quickly, reconstructs corrupted images more accurately, and generates samples with a high estimated log-likelihood. Lastly, we note that, interestingly, if an asymmetric version of VPF exists, the weight updates directly explain experimental results in Spike-Timing-Dependent Plasticity (STDP). 
\end{abstract}

\section{Introduction}
\label{intro}

With the immense success of deep learning in machine learning and artificial intelligence, there has been significant interest in determining the extent to which it explains learning in biological neural networks. For example, Bengio, et al. (2015) have pointed out several challenges in finding biologically plausible implementations of backpropagation, such as the need for precisely-clocked linear feedback paths that have exact knowledge of the weights and the derivatives involved in the feedforward paths. Moreover, backpropagation requires output targets, and passes real-valued signals rather than binary-valued spikes between neurons \cite{bengio2015towards,rumelhart1988learning}. Contrastive divergence, on the other hand, trains a binary generative model, but it involves Gibbs sampling to generate data confabulations for the computation of a negative gradient \cite{hinton2002training}. It is unclear how this may be implemented biologically.

Minimally, the update rules for a biologically plausible learning algorithm should be local. More precisely, the change in the weight of a synapse should depend only on information that is directly accessible by the synapse, such as the states and the firing rates of the adjacent neurons. Local rules are also beneficial for machine computation. When the weight updates are not local, such as in the case of backpropagation, it may be necessary to store all the weights and activities of a massive neural network in the memory of the CPU or GPU.  Schemes which bypass the need to store all the weights often pay the price with slower convergence to the optimal solution \cite{jaderberg2016decoupled}.

Spike-Timing-Dependent Plasticity (STDP) is another attribute of biological networks \cite{markram1995action,gerstner1996neuronal,bi1998synaptic,xie2000spike}. This rule states that the change in the weights increases as the time interval between the pre-synaptic and post-synaptic spike decreases, and the sign of the change depends on the temporal ordering of the two spikes. If the pre-synaptic neuron spikes before the post-synaptic neuron, then the weight of the synapse increases, almost as if to increase the probability of the post-synaptic spike at the next pre-synaptic spike. If the order of the spikes is switched, then the synaptic weight decreases. Bengio, et al. (2015) proposed a novel weight update and showed via simulation that it gives rise to STDP. This update rule was then justified using a new learning framework and energy-based objective function.

The main contribution of this paper is the proposal of a new learning algorithm, Variational Probability Flow (VPF), that is biologically plausible in the following ways.
\begin{enumerate}
\item The update rule for a given weight $w_{ij}$ is local. It only depends on the binary-valued states $y_i, y_j$ and the real-valued firing rates $\delta_i, \delta_j$ of the adjacent neurons.
\item An asymmetric version of the update explains STDP. 
\item The model neurons are binary-valued, not real-valued.
\item The model is generative so output targets are not required.
\item The updates avoid feedback paths and confabulations.
\item The updates allow intra-layer connections to be trained.
\end{enumerate}
In fact, the asymmetric update that we discovered is
$$
\Delta w_{ij} \,\,\propto\,\, y_i \delta_j
$$
where $y_i$ is the pre-synaptic state and $\delta_j$ is the post-synaptic firing rate. This form is different from the update proposed in \cite{bengio2015towards} and \cite{hinton2007backpropagation}, where $\Delta w_{ij}$ depends not on the rate $\delta_j$ of post-synaptic transitions but on the \emph{rate of change} of the integrated post-synaptic activity.

As its name suggests, VPF is a variational extension of minimum probability flow \cite{sohl2011new,sohl2009minimum}. We derive a variational objective function for training Boltzmann machines (BMs) with observed and hidden variables, and prove that it is an upper bound to  the probability flow of the observed variables. We then outline a variational expectation-maximization (EM) algorithm that minimizes this objective \cite{neal1998view}. 
Experiments with restricted Boltzmann machines (RBMs) show that VPF is able to quickly learn features which resemble pen strokes and reconstruct corrupted digits more accurately than previous methods. 
For deep Boltzmann machines (DBMs), we also propose a new strategy that does not require greedy layer-wise pre-training \cite{hinton2006reducing}. 
Moreover, VPF applies directly to BMs with intra-layer connections which significantly improve the model's ability to learn sparse representations. When applied to MNIST and Fashion MNIST data, experiments show that VPF is able to generate realistic samples with estimated log-likelihoods similar to that of generative adversarial networks (GANs). In future work, we will explore how the local learning rules may be exploited to design model-parallel algorithms that run efficiently on GPUs. 

\section{Background}
\label{background}

\subsection{Boltzmann Machines}

BMs are energy-based probabilistic models \cite{ackley1985learning}. Given an undirected graph $\mathcal{G} = (\mathcal{V}, \mathcal{E})$ without loops or multiple edges, we associate a binary random variable $x_i$ and a real \emph{bias} $b_i$ to each vertex $i \in \mathcal{V}$, and a real \emph{weight} $w_{ij}$ to each edge $ij \in \mathcal{E}$. Let the probability of each vector $\bm{x}=(x_i)_{i \in \mathcal{V}}$ be $p(\bm{x}) \propto \exp (- \frac{1}{T} \text{Energy}(\bm{x}))$
where $\text{Energy}(\bm{x}) = - \sum_{ij \in \mathcal{E}} w_{ij}x_ix_j - \sum_{i \in \mathcal{V}} b_i x_i$, and $T$ is the temperature. For simplicity, we fix $T=1$ in this paper. 

In statistical learning, it is useful to designate some of the variables $x_i$ to be observed and the others to be hidden. The probability of each observed state vector is obtained by marginalizing over all the hidden states. In this paper, we use multilayer networks to refer to BMs with a layered structure, where the connections are either within the layers or between consecutive layers. 
For instance, DBM is a special case with layers $\bm{h}_0, \bm{h}_1, \ldots, \bm{h}_\ell$ where the first layer $\bm{h}_0$ is observed while the others are hidden. RBM is special cases of DBMs with $\ell = 1$. 
In these underlying graph, there are only edges between consecutive layers $\bm{h}_i$ and  $\bm{h}_{i+1}$. 
DBMs with intra-layer connections are studied as well. 

\subsection{Minimum Probability Flow}
\label{mpf}

Maximum likelihood estimation can be intractable for many important classes of statistical models. To overcome this problem, Sohl-Dickstein, et el. (2011) designed a new learning objective known as Minimum Probability Flow (MPF) for finite state models which may be sampled effectively using a suitable continuous-time Markov chain (CTMC) \cite{sohl2011new,sohl2009minimum}. Let $\bm{p}^{(t)}(\theta)$ be a CTMC parameterized by $\theta$ such that $p^{(t)}(\bm{x};\theta)$ is the probability of state $\bm{x}$ at time $t$ in the Markov chain. Let $\bm{p}^{(0)}(\theta) = \bm{p}^{(0)}$ denote the empirical distribution of the data set $\mathcal{D} = \{ \bm{x}^{(1)}, \ldots, \bm{x}^{(N)} \}$ of binary vectors, and we assume the stationary distribution $\bm{p}^{(\infty)}(\theta)$ is a distribution in the finite state model that we hope to fit to the data.

Using this notation, maximum likelihood estimation is equivalent to minimizing the Kullback-Leibler divergence $D_{\text{KL}} (\bm{p}^{(0)} \| \bm{p}^{(\infty)} (\theta))$ over $\theta$. On the contrary, minimum probability flow computes
\begin{equation*}
\min_\theta \mathcal{K}(\theta), \quad  \mathcal{K}(\theta) := D_{\text{KL}} (\bm{p}^{(0)} \| \bm{p}^{(\varepsilon)} (\theta)),
\end{equation*}
for infinitesimal time $\varepsilon > 0$. Given an energy-based model $p^{(\infty)}( \bm{x};\theta) \propto \exp(-\text{Energy}(\bm{x};\theta))$ where $\text{Energy}(\bm{x};\theta)$ is the energy of state $\bm{x}$, we may construct a CTMC $\bm{p}^{(t)}(\theta) = \exp(\bm{\Gamma}(\theta) t) \bm{p}^{(0)}$ that attains this stationary distribution by defining its transition rate matrix $\bm{\Gamma} = (\Gamma_{xy})$ to be
$$
\Gamma_{xy} = g_{xy} \exp \left(\frac{1}{2}\text{Energy}(\bm{y};\theta) - \frac{1}{2}\text{Energy}(\bm{x};\theta)\right),   \bm{x} \neq \bm{y},
$$
where $\Gamma_{xx} = -\sum_{\bm{y} \neq \bm{x}} \Gamma_{xy}$ and each $g_{xy} \in \{0,1\}$ is some choice of connectivity between states $\bm{x}$ and $\bm{y}$. In this case, one can show that for small $\varepsilon > 0$, 
{\small \begin{align*}
\mathcal{K}(\theta)  & = \frac{\varepsilon}{N} \!\!\! \sum_{\bm{x} \notin \mathcal{D},\bm{y} \in \mathcal{D}} \!\!\! \Gamma_{xy}   \\
 & = \frac{\varepsilon}{N}\!\!\! \sum_{\bm{x} \notin \mathcal{D},\bm{y} \in \mathcal{D}} \!\!\!g_{xy} \exp\left( \!\frac{1}{2} \text{Energy}(\bm{y};\theta) - \frac{1}{2}\text{Energy}(\bm{x};\theta)\!\right)\!. 
\end{align*}}

Therefore, MPF is a learning framework for energy-based models that avoids computing the intractable partition function and its derivatives.

\section{Variational Probability Flow}
\label{vpfl}

In this section, we derive VPF for hidden variable models by applying variational principles to probability flow.

\subsection{Fully-Observed Boltzmann Machines}
\label{VMPF}

We first apply MPF to the training of fully-observed Boltzmann machines, by defining the connectivity $g_{xy}$ to be $1$ if and only if the states $\bm{x}$ and $\bm{y}$ are one-hop neighbors, i.e. they differ only by one bit. Given training data 
$\mathcal{D} = \{\bm{y}^{(1)}, \ldots, \bm{y}^{(N)} \}$, the MPF objective function becomes
\begin{equation*}
	\mathcal{K}(\theta) = \frac{{\varepsilon}}{N} \sum_{k=1}^{N}\mathcal{K}^{(k)}(\theta), \quad \mathcal{K}^{(k)}(\theta) =\sum_{j=1}^{|\mathcal{V}|} \delta_j^{(k)},
\end{equation*}
where for convenience we define $$ \alpha_j^{(k)} = \tfrac{1}{2} - y^{(k)}_j, z_j^{(k)} = \textstyle \sum_{i\neq j} w_{ij}y^{(k)}_i +b_j,$$ $$\delta_j^{(k)} = \exp(\alpha_j^{(k)}z_j^{(k)}).$$
To update the parameters, we use the negative gradients
\begin{equation}
\label{updateb}
\Delta b_i \propto - \alpha_i^{(k)} \delta_i^{(k)}, 
\end{equation}
\begin{equation}
\label{updatew}
\Delta w_{ij} \propto -\,\, y^{(k)}_j \alpha_i^{(k)} \delta_i^{(k)} - y^{(k)}_i \alpha_j^{(k)} \delta_j^{(k)}.
\end{equation}
We note that $\delta_j^{(k)}$ is the transition rate $\Gamma_{xy}$ where $\bm{y} = \bm{y}^{(k)}$ and $\bm{x}$ is the one-hop neighbor that differs in the $j$-th bit. In other words, it is the firing rate of a Poisson process that determines the flipping of the state of the $j$-th vertex. Please refer to the supplementary material for detailed proofs. 

\subsection{VPF Objective Function}

Consider a Boltzmann machine with observed and hidden variables $\bm{x}, \bm{h}$. Let $\tilde{\mathcal{D}} =\{\bm{x}^{(1)}, \ldots, \bm{x}^{(N)} \}$ be \textsc{iid} samples of $\bm{x}$, and $\tilde{\bm{p}}^{(0)}$ be the corresponding empirical distribution. Let $p^{(t)}(\bm{x},\bm{h};\theta)$ be the associated CTMC. Our goal is to minimize the marginal probability flow 
$D_{\text{KL}} (\tilde{\bm{p}}^{(0)} \|\tilde{\bm{p}}^{(\varepsilon)} (\theta))$ where
{$\tilde{p}^{(t)} (\theta)$} represents the marginal probability $p^{(t)}(\bm{x};\theta)$. Let the VPF objective function be
$$
\mathcal{L}^{(t)}(q,\theta)= D_{\text{KL}} (q(\bm{h},\bm{x}) \|p^{(t)} (\bm{h},\bm{x};\theta)) 
$$
where $q := q(\bm{h}|\bm{x})$ is some distribution over the hidden variables given the observed variables, and $q(\bm{x},\bm{h})$ denotes $q(\bm{h}|\bm{x}) \tilde{p}^{(0)}(\bm{x})$. As shown in the supplementary material,
\begin{align}\label{eq:kullback-trio}
\nonumber \mathcal{L}^{(t)}(q,\theta) &= \mathbb{E}_{\bm{x}\sim \tilde{p}^{(0)}(\bm{x})}\left[ D_{\text{KL}} (q(\bm{h}|\bm{x}) \|p^{(t)} (\bm{h}|\bm{x};\theta)) \right] \nonumber + \\
& \quad \quad D_{\text{KL}} (\tilde{\bm{p}}^{(0)} \|\tilde{\bm{p}}^{(t)} (\theta)).
\end{align}
Therefore, if we hold $\theta$ constant and minimize $\mathcal{L}^{(t)}(q,\theta)$ over all conditional distributions $q$, then the minimum occurs when $q(\bm{h}|\bm{x}) = p^{(t)} (\bm{h}|\bm{x};\theta)$. However, it is not possible to compute the latter distribution unless the joint distribution $p^{(t)}(\bm{x},\bm{h})$ at time $t=0$ is known. We will therefore approximate it by assuming that the CTMC is close to the stationary distribution $p^{(\infty)}(\bm{x},\bm{h})$, and select $q(\bm{h}|\bm{x}) = p^{(\infty)}(\bm{h}|\bm{x};\theta)$. This approximation implies that $\mathcal{L}^{(t)}(q,\theta)$ is an upper bound to 
$D_{\text{KL}} (\tilde{\bm{p}}^{(0)} \|\tilde{\bm{p}}^{(t)} (\theta))$, and the bound improves as $p^{(t)} (\bm{h}|\bm{x};\theta)$ tends to the model $p^{(\infty)}(\bm{h}|\bm{x};\theta)$. 

If it is biologically desirable to perform the above optimizations while constraining the weights $w_{ij}$ to a bounded space, one could do so by adding a weight decay term to the objective functions, e.g., $\min_{q,\theta} \mathcal{L}^{(t)}(q,\theta) + \lambda \sum_{ij} w_{ij}^{2}$. 

\subsection{VPF Learning Algorithm}

The objective function is optimized with a variational EM algorithm. The E-step and M-step are repeated until convergence of the parameters $\theta$ of the Boltzmann machine.

\noindent \textbf{E-step.} We set $q(\bm{h},\bm{x}) = p^{(\infty)}(\bm{h}|\bm{x};\theta) \,\tilde{p}^{(0)}(\bm{x})$ and compute $\mathcal{L}^{(\varepsilon)}(q,\theta)$ given this choice. We further approximate 
\begin{align} \label{estep}
\mathcal{L}^{(\varepsilon)}(q,\theta) & = \int q(\bm{x},\bm{h}) \log q(\bm{x},\bm{h})\, d\bm{x}d\bm{h}  -   \nonumber \\ 
& \quad \int q(\bm{x},\bm{h}) \log {p^{(\varepsilon)} (\bm{h},\bm{x};\theta)} \,d\bm{x}d\bm{h}
\end{align}
by substituting the latter integral with an average over samples from $q(\bm{x},\bm{h})$. By optimizing the resulting stochastic objective $\mathcal{S}(\theta)$, the parameters will still converge to the same limit as that of $\mathcal{L}^{(\varepsilon)}(q,\theta)$ \cite{gal2015dropout}.


\begin{algorithm}[!htb]{}
\caption{VPF for training a BM.}
\label{VPFDBM}
\begin{algorithmic}[1]
\STATE  $\bm{x} \gets$ training data
\STATE  $\bm{w},\bm{b} \gets $ random initialization
\WHILE{$ (\bm{w}, \bm{b})$ not converged}
	\FOR  {$i = 1$ to $\ell$ }
   		\STATE $\bm{h}_i \gets $ Bernoulli sample of $\bm{h}_i$ given $\bm{h}_{i-1}$ 
		\IF {intra-layer connections exist}
		\STATE Asynchronously update $\bm{h}_i$ following Eq. \ref{asyn}
		\ENDIF 
    \ENDFOR
	\STATE $\bm{y} \gets (\bm{x},\bm{h})$ 
	\FOR {minibatch $\bm{y}_m$ in $\bm{y}$ }
		\STATE $\Delta \bm{w}, \Delta \bm{b} \gets $ MPF updates at $\bm{w}, \bm{b}$ for $\bm{y}_m$
		\STATE Update parameters $\bm{w},\bm{b}$ with $\Delta \bm{w}, \Delta \bm{b}$ using Adam optimizer
	\ENDFOR
\ENDWHILE
\RETURN $\bm{w},\bm{b}$
\end{algorithmic}
\end{algorithm}

\begin{algorithm}[!htb]{}
\caption{Generating confabulations from a given BM.}
\label{DBM-confab}
\begin{algorithmic}[1]
\STATE  $\bm{h}_\ell \gets$ random initialization
	\FOR  {$i = \ell$ to $1$ }
\FOR {$j =1$ to $r$}
   	\STATE $\bm{h}_{i-1} \gets $ Bernoulli sample of $\bm{h}_{i-1}$ given $\bm{h}_{i}$
   	\STATE $\bm{h}_{i} \gets $ Bernoulli sample of $\bm{h}_{i}$ given $\bm{h}_{i-1}$
	\IF {intra-layer connections exist}
		\STATE Asynchronously update $\bm{h}_{i}$ following Eq. \ref{asyn}
	\ENDIF
\ENDFOR
\ENDFOR
\RETURN Bernoulli probability of $\bm{h}_{0}$ given $\bm{h}_{1}$
\end{algorithmic}
\end{algorithm}

To sample from the conditional distribution $p^{(\infty)}(\bm{h}|\bm{x};\theta)$ of the Boltzmann machine, the hidden states could be initialized randomly before applying Gibbs sampling. For the RBM and the DBM, we will discuss approximate sampling schemes in next section. From our experiments, we observe that it is enough to generate just one sample to achieve good convergence of the learning algorithm. 

\noindent \textbf{M-step.} We minimize the stochastic objective $\mathcal{S}(\theta)$. In the E-step, we obtained samples of $\bm{h}$ given each data point $\bm{x}$, giving us the set $\mathcal{D} = \{(\bm{x}^{(1)},\bm{h}^{(1)}), \ldots, (\bm{x}^{(N)},\bm{h}^{(N)})\}$. The objective $\mathcal{S}(\theta)$ is precisely the probability flow of a fully-observed Boltzmann machine with data $\mathcal{D}$, so Eq. \ref{updateb}, \ref{updatew} may be applied to update the parameter $\theta$.

\subsection{VPF for Multilayer Networks}
\label{vpfdeep}
The variational EM algorithm allows us to train any BMs with hidden variables. 
In this section, we discuss strategies for improving its performance on BMs where the hidden variables 
have a layered structure.

\noindent \textbf{RBM.} For RBMs, the hidden variables $\bm{h}$ are conditionally independent given the observed variables $\bm{x}$ with 
$$
p^{(\infty)}(h_j=1|\bm{x};\bm{w},\bm{b}) = \text{sigmoid}(\textstyle \sum_{i} w_{ij}x_i + b_j).
$$
These Bernoulli probabilities can be computed and sampled efficiently on a machine. The pseudocode for applying VPF to RBMs is given in Algorithm. \ref{VPFDBM} with $\ell=1$. Stochastic gradient descent (SGD) is exploited during training but with a minor change. Traditionally, when SGD is performed in deep learning, the parameters are updated after each mini-batch. In VPF, at the start of each epoch, we perform the E-step for the entire data set, sampling the hidden states for each data point using the parameters attained at the end of the last epoch. The observed-hidden pairs are then partitioned into mini-batches for the M-step. This procedure seems to promote convergence to better local minima. 

\noindent \textbf{DBM.} For DBMs with $\ell \geq 2$ hidden layers, the hidden variables $\bm{h}$ are no longer conditionally independent given the observed variables $\bm{x}$. One strategy for training a DBM is to train each pair $(\bm{h}_{i-1}, \bm{h}_{i})$ of layers as an RBM, starting from $\bm{h}_{0}$ and generating data for $\bm{h}_{i}$ after $(\bm{h}_{i-1}, \bm{h}_{i})$ has been trained. Here, we propose a different strategy that does not require greedy layer-wise training. We approximate the conditional distribution $p^{(\infty)}(\bm{h}|\bm{x};\bm{w},\bm{b})$ by computing Bernoulli probabilities for $\bm{h}_{i}$ using only the weighted inputs from $\bm{h}_{i-1}$ while zeroing out the inputs from $\bm{h}_{i+1}$. We sample $\bm{h}_i$ from these probabilities before proceeding in the same way to sample $\bm{h}_{i+1}$. The generated hidden states are then used for the M-step where the weights for every layer are updated at the same time. The pseudocode for training a DBM is shown in Algorithm. \ref{VPFDBM}. 

To generate confabulations from a given DBM, we start at the top layer with a binary vector $\bm{h}_\ell$ that was picked uniformly or from some priors. We perform $r$ Gibbs sampling steps, alternating between $\bm{h}_\ell$ and $\bm{h}_{\ell-1}$. In our experiments, it seems that $r=5$ was enough to ensure good mixing. The final state of $\bm{h}_{\ell-1}$ is then used to initialize the next $r$ Gibbs sampling steps for generating $\bm{h}_{\ell-2}$, and so on. The pseudocode for this process is given in Algorithm. \ref{DBM-confab}.

\noindent \textbf{DBM with Intra-layer Connections.} 
We introduce how to employ VPF to train DBMs with intra-layer connections, which  
can be viewed as a stack of RBMs but with connections between the hidden nodes, 
see Fig.\ref{generation}.(a). To train these generalized DBMs, we take an additional step of performing asynchronous Gibbs updates for the hidden units $\bm{h}_i$ given the values of $\bm{h}_{i-1}$, $\bm{h}_i$ in the previous update. 
More specifically, we update the neurons in $\bm{h}_i$ one by one in a fixed order, sampling the $j$-th neuron $h_{ij}$ from the probability
\begin{equation}
\label{asyn}
p_{ij} = \text{sigmoid}(\textstyle \sum_{k\neq j} w_{kj}h_{ik} + \sum_{k} w'_{kj}h_{(i-1)k} + b_j).
\end{equation}
where $\bm{w}, \bm{w'}$ denote the intra- and inter-layer weights respectively. Updates for the intra-layer weights also follow the local rules Eq. \ref{updateb}, \ref{updatew}. 
The confabulation-generating procedure has an additional step involving intra-layer Gibbs updates.
The modifications are indicated in Algorithm. \ref{VPFDBM}, \ref{DBM-confab}.

As with traditional associative memories, intra-layer connections help the model to learn rich energy landscapes and sparse distributed representations which are otherwise impossible for disconnected layers with factorial distributions. By embracing them in VPF, we hope to learn, from the observed data, hidden priors with greater explanatory power. Usually, one assumes the absence of connections within each layer so as to derive tractable training algorithms. VPF overcomes this limitation, through the combination of probability flow with variational techniques. In our experiments, we explore the effect that these intra-layer connections have on the model's performance in generative tasks.

\section{Spike-Timing-Dependent Plasticity}
\label{stdp}

STDP is a phenomenon observed in biological neural networks, but little is understood about how it contributes to learning. In this section, we show how VPF could give rise to STDP. First, we present a simplified model of learning in biological neural networks. Each neuron $y_i$ in our model are either in an excited state $y_i=1$ or in a resting state $y_i=0$ \cite{amari1977neural}. The time it takes for an excited neuron to transit to a resting state follows an exponential distribution with rate $\delta_i = \exp(- z_j)$, while the time for a resting neuron to become excited occurs with rate $\delta_i = \exp(+z_j)$. The synaptic weight $w_{ij}$ is updated if the post-synaptic neuron $j$ transits and if the pre-synaptic neuron $i$ is excited. The change in value is proportional to
\begin{align}\label{eq:stdp-rule}
\Delta w_{ij} \propto - y_i \alpha_j \delta_j,
\end{align}
where $\delta_j$ is the post-synaptic rate before the transition, $y_j$ is the post-synaptic state after the transition, and $\alpha_j = 1/2-y_j= \pm 1/2$ only affects the sign of the update. This update rule is local, in the sense that it depends only on the states $y_i, y_j$ and the firing rates $\delta_i, \delta_j$ of the adjacent neurons. The weights $w_{ij}=w_{ji}$ are symmetric in BMs \cite{scellier2017equilibrium}, so we combine the updates for $\Delta w_{ij}$ and $\Delta w_{ji}$ from Eq. \ref{eq:stdp-rule}, which gives the VPF update Eq. \ref{updatew}.

We now show that update Eq. \ref{eq:stdp-rule} explains STDP.  Suppose that we have a pre-synaptic spike before a post-synaptic spike. Specifically, at time $t=0$, we have $y_i(t) = 1$ and $y_j(t)=0$, and at some $t = \varepsilon$, we have $y_j(t)=1$. There are two cases to consider: either the pre-synaptic neuron is still excited and $y_i(\varepsilon)=1$, or the pre-synaptic neuron has returned to resting state and $y_i(\varepsilon)=0$. The first case occurs with probability $\exp(-\delta_i \varepsilon)$ and the weight update is $\Delta w_{ij} \propto \delta_j$. Here, $\delta_j = 1/\varepsilon$ because the expected time to transition is the inverse of the transition rate. The second case occurs with probability $1-\exp(-\delta_i \varepsilon)$, but the weight update is $0$ because $y_i(\varepsilon)=0$. Thus, the expected weight change is 
$$\Delta w_{ij} \propto (1/\varepsilon) \exp(-\delta_i \varepsilon).
$$

Suppose instead that the pre-synaptic spike happens after a post-synaptic spike: $y_i(0) = 0, y_j(0)=1$, and  $y_i(\varepsilon)=1$. Again, there are two cases: either the post-synaptic neuron is still excited and $y_j(\varepsilon)=1$, or the post-synaptic neuron has returned to resting state and $y_j(\varepsilon)=0$. The first case occurs with probability $\exp(-\delta_j \varepsilon)$. After the pre-synaptic spike, there is no immediate weight change, until the first post-synaptic flip to state $0$ takes place. The update is $\Delta w_{ij} \propto - \delta_j$ for this first flip. The second case happens with probability $1-\exp(-\delta_j \varepsilon)$. However, in the absence of stimuli, the post-synaptic neuron does not become excited again before the pre-synaptic neuron returns to rest. Hence, the update is 0. Overall, the average weight change is 
$$\Delta w_{ij} \propto -\delta_j \exp(-\delta_j \varepsilon).
$$

\begin{figure}[t]
  \centering
  \subfigure[STDP function]{\includegraphics[scale=0.27]{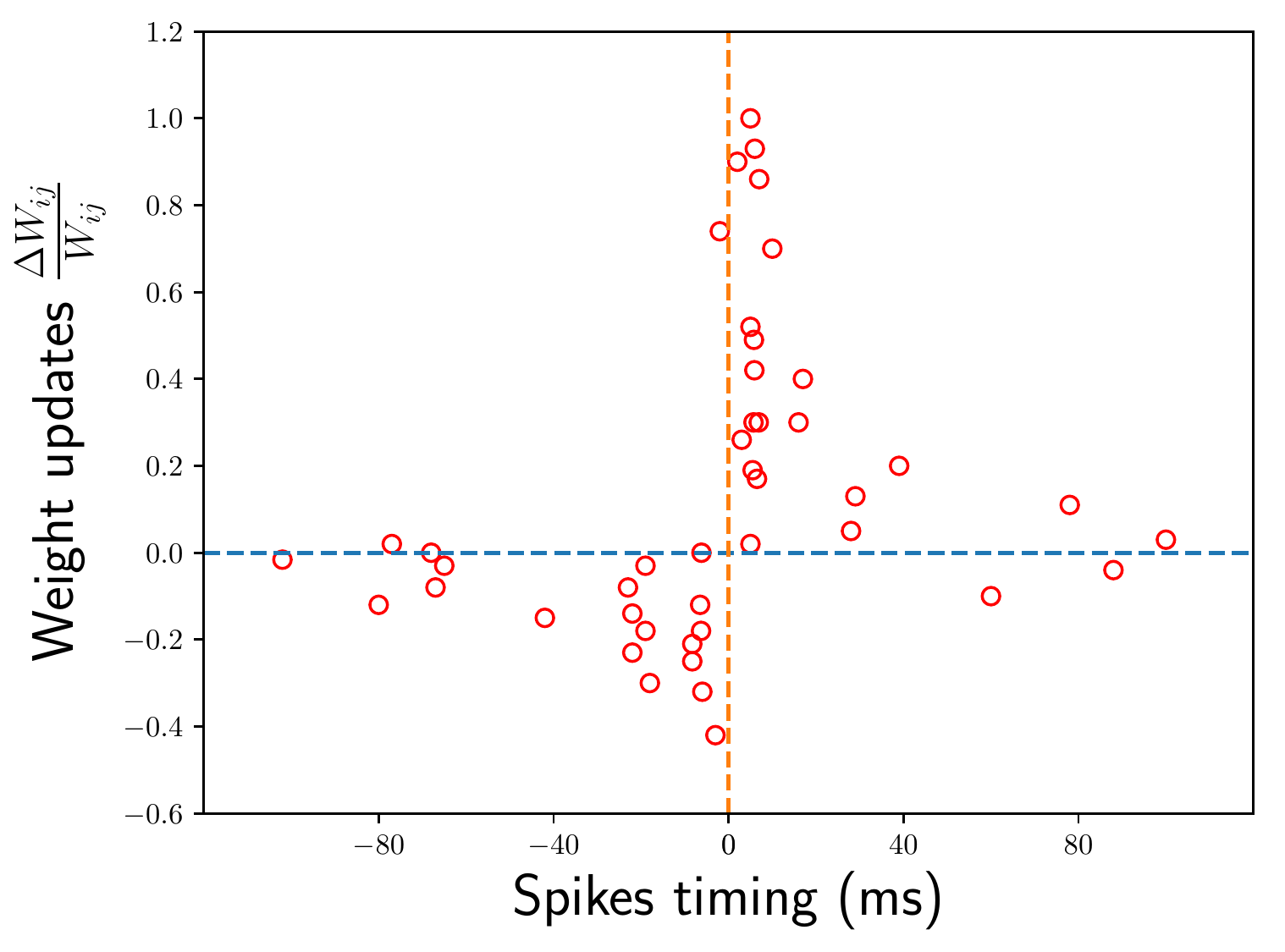}} 
  \subfigure[Weight updates under VPF]{\includegraphics[scale=0.27]{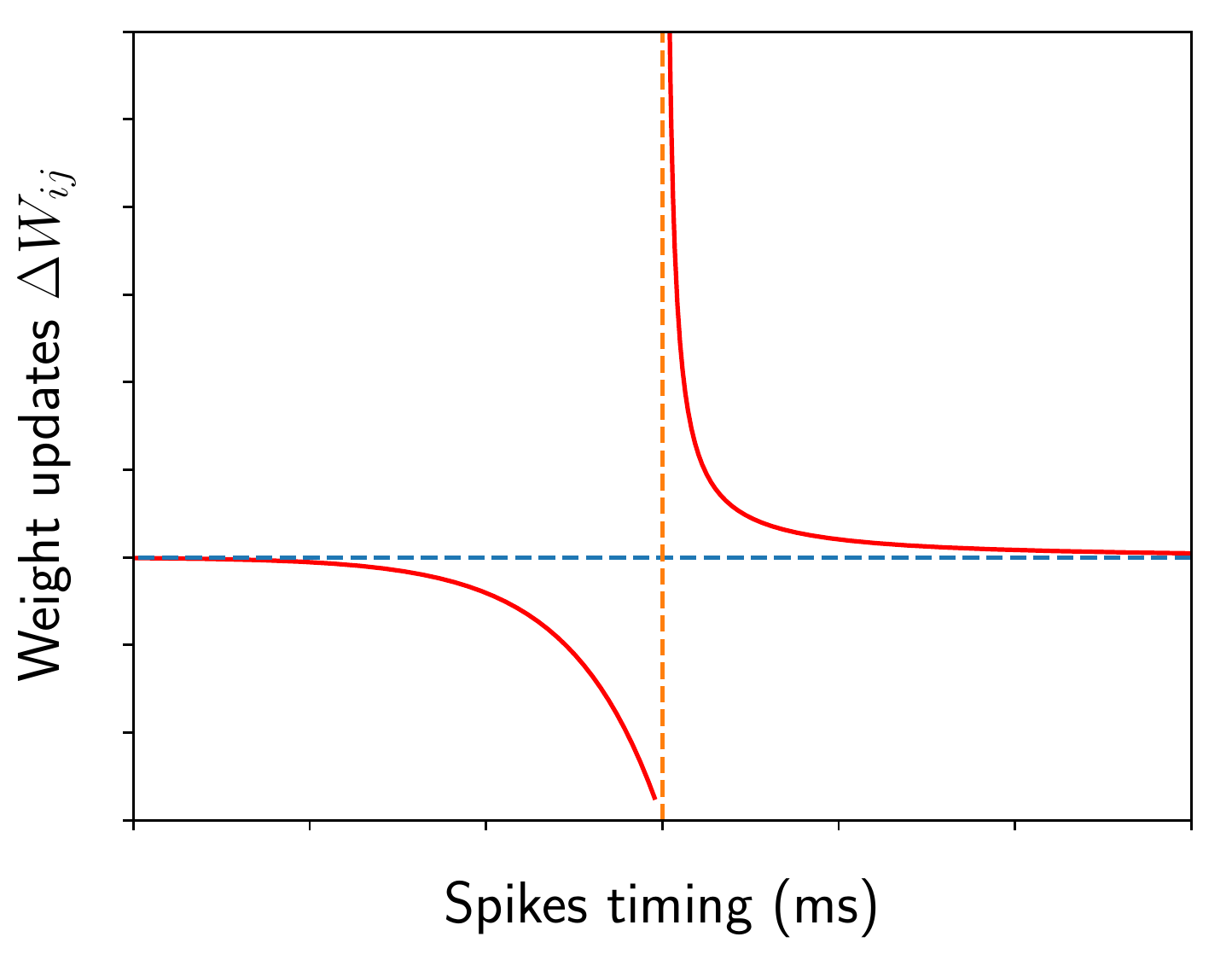}} 
   \caption{ {\bf(a)}. The trend in STDP, reproduced from \cite{bi1998synaptic}. {\bf(b)}. A visualization of how expected weight updates are inversely proportional to time intervals between the spikes, as predicted by VPF.}
  \label{stdpfunction}
\end{figure}

Plotting the two average weight updates derived above in Fig.\ref{stdpfunction}, we see a good fit with the experimental measurements for STDP. In future, we hope to find a statistical model that, unlike BMs, does not require symmetry, and whose update rule under VPF is precisely given by Eq. \ref{eq:stdp-rule}.

\section{Experiments}
\label{experiments}

We conducted experiments on MNIST digits and Fashion MNIST images. MNIST is binarized by thresholding with $0.5$, and Fashion MNIST is scaled to $[0, 1]$ before thresholding. The Adam optimizer is used during optimization with default values for the parameters, e.g., $\beta_1 = 0.9$, $\beta_2 = 0.999$ \cite{kingma2014adam}. The learning rate is $\eta = 0.001$. The weight decay parameter $\lambda$ is set to $0.0001$ for all the RBMs and DBMs. We used a mini-batch size of $M = 40$, and the code is implemented with Theano \cite{team2016theano}. More details can be found in the supplementary material. 

\subsection{Restricted Boltzmann Machines}
\begin{figure}[t]
  \centering
  \subfigure[RBM(100)]{\includegraphics[scale=0.52]{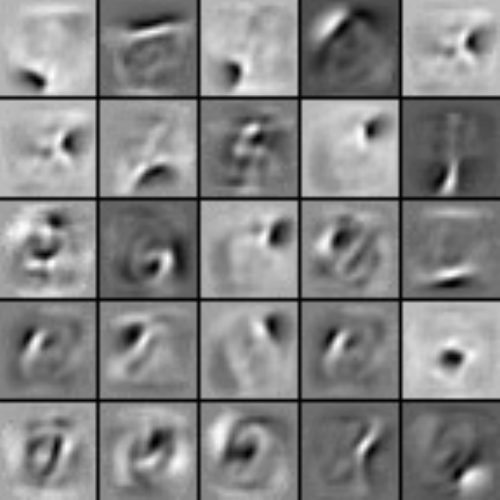}} 
  \subfigure[RBM(196)]{\includegraphics[scale=0.52]{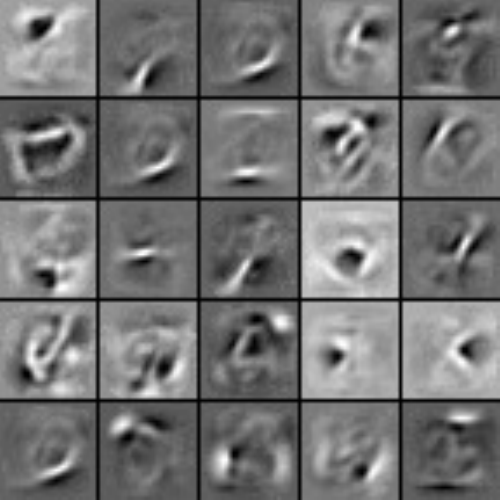}} 
  \subfigure[RBM(400)]{\includegraphics[scale=0.52]{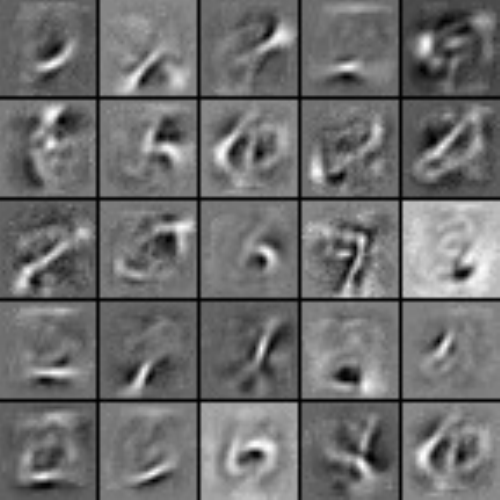}}
   \caption{Randomly selected filters learned from MNIST. }
  \label{filters}
\end{figure}

\begin{figure}[t]
\centering
	\subfigure[]{\includegraphics[scale=0.173]{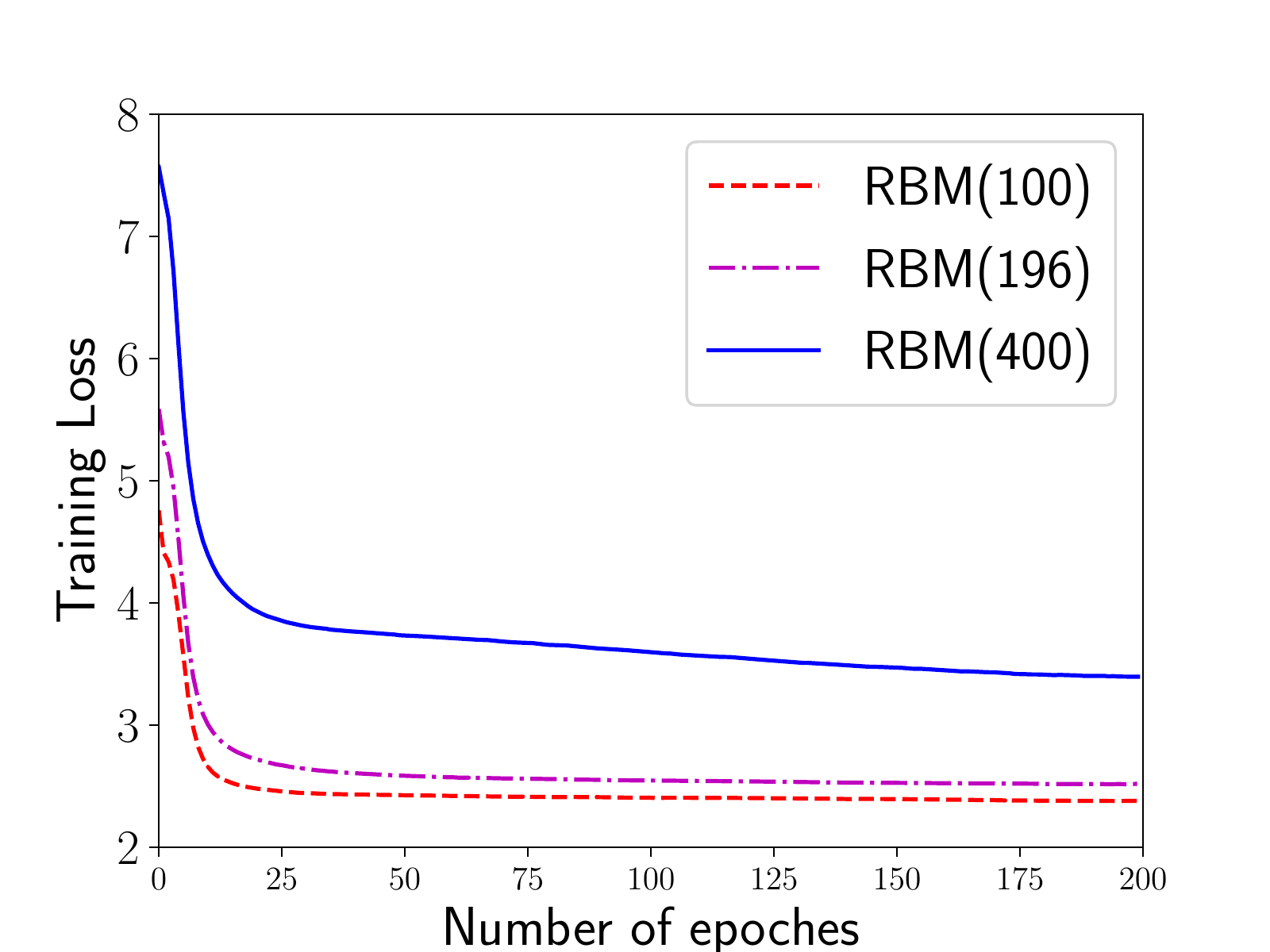}} 
	\subfigure[]{\includegraphics[scale=0.173]{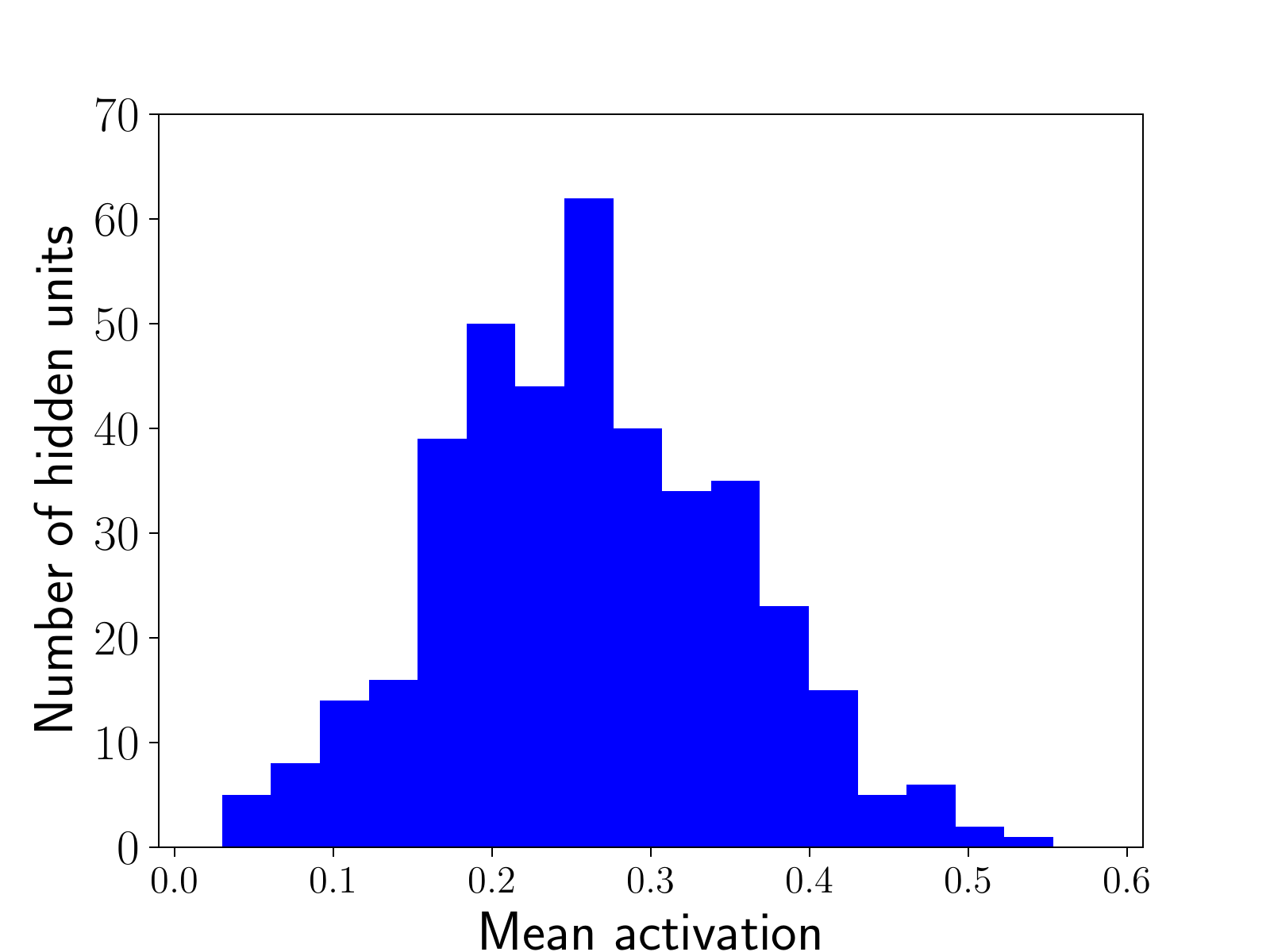}}
       \subfigure[]{\includegraphics[scale=0.161]{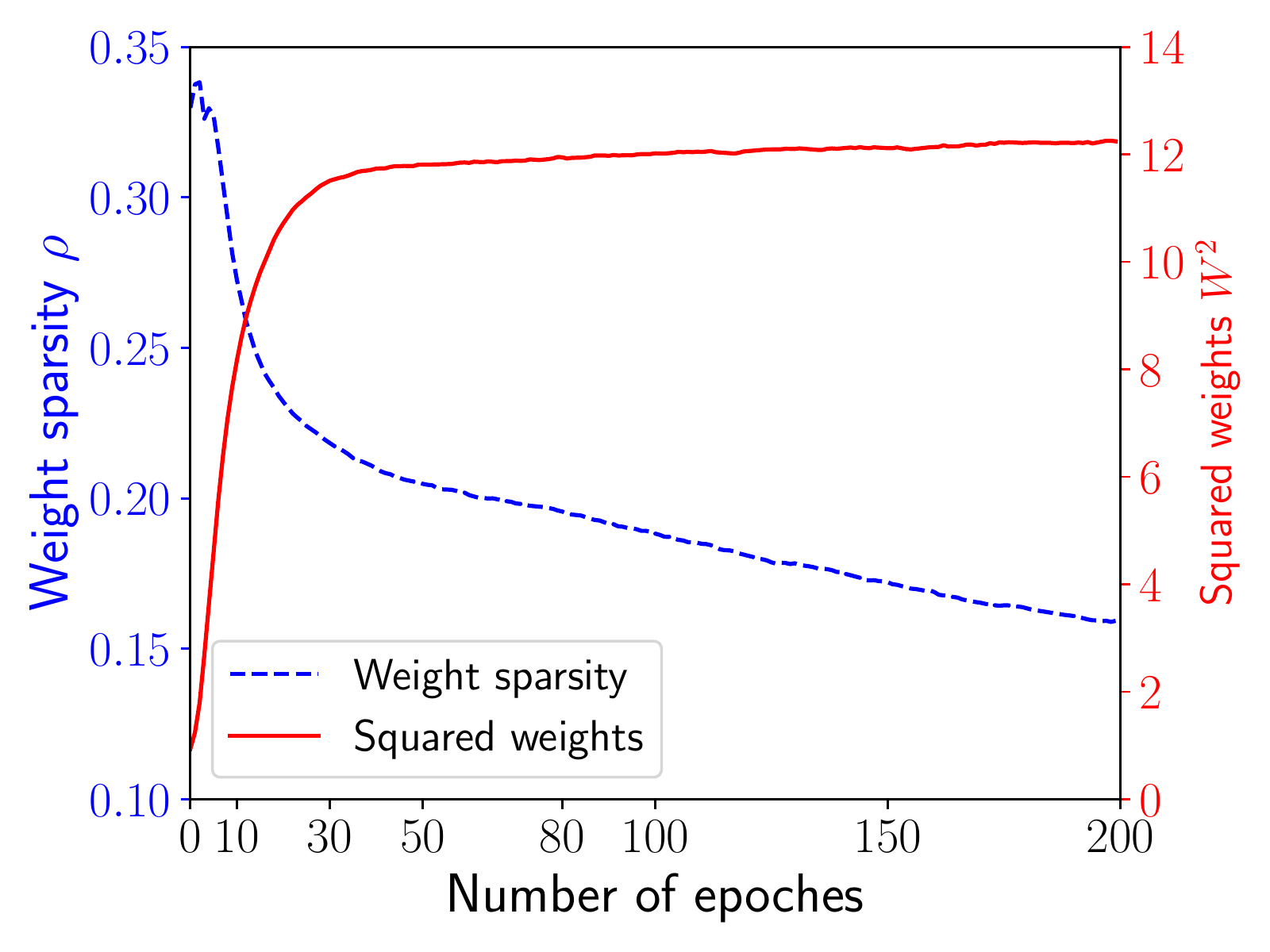}} 
\caption{ {\bf(a)}.  The training loss of different RBMs. {\bf(b)}. Histogram of mean activations of RBM(400). 
{\bf(c)}. Evolution of the weight sparsity and squared weight of RBM(400). }
\label{activations}
\end{figure}

In the first experiment, we trained three RBMs: RBM(100), RBM(196), RBM(400), using Algorithm. \ref{VPFDBM} with $\ell = 1$,  
where 100, 196, 400 denote the number of hidden units. Fig. \ref{filters} shows 25 randomly selected weight filters of the RBMs. 
All of the models learned reasonable filters (pen strokes), which are sharper in RBM(196) and RBM(400). 
We also noticed that the training objective converged quickly. As shown in Fig.\ref{activations}.(a), all the curves flattened out in less than 50 epochs. 

To verify that VPF learns a distributed representation, we plotted a histogram of the mean activations of hidden units, which we averaged over 10,000 training examples. As shown in Fig.\ref{activations}.(b), majority of the hidden units have mean activations between $0.1$ and $0.4$. The overall mean activation is $0.26$, which is reasonably sparse for MNIST data. We also tracked two other metrics, the weight sparsity $\rho$ and the squared weight $w^2$, which are given by
$$ \rho = \frac{1}{|\bm{x}||\bm{h}|} \sum_j[  (\sum_i w_{ij}^2)^{2} / \sum_i w_{ij}^{4}], \quad w^{2} = \frac{1}{|\bm{h}|} \sum_{ij} w_{ij}^{2},$$
where $|\bm{x}|, |\bm{h}|$ are the dimensions of $\bm{x}, \bm{h}$, and $w_{ij}$ is the weight from visible unit $i$ to hidden unit $j$. The weight sparsity $\rho$ is a rough measure of the number of nonzero weight parameters. It decreased to $0.15$ during training, as shown in Fig.\ref{activations}.(c), which suggests that many weights are near zero. The remaining nonzero weights grew in value during training, as may be seen in the increasing squared weight $w^2$ in Fig.\ref{activations}.(c) and the high-contrast features in Fig. \ref{filters}. These results demonstrate that RBMs trained with VPF operate in a compositional phase with distributed representations \cite{tubiana2017emergence}. 

\subsection{Image Reconstruction with RBMs}
We explored the expressiveness of  the VPF model by comparing RBM(196) to RBMs trained with CD and 
persistent contrastive divergence (PCD) in terms of the ability to reconstruct corrupted parts of MNIST digits.  
To reconstruct the images, we run Markov chains starting from the corrupted images and performed 1000 
Gibbs transitions for CD and PCD to make them mix well. 
For RBM(196), we found that 2 Gibbs transitions were enough for good results. Codes for CD and PCD were derived from \cite{team2016theano}. 

From Fig. \ref{reconstruction} we can see that RBM(196) is able to reconstruct the initially corrupted 
pixels under 4 different corruption patterns. The reconstruction errors were measured as the $\mathcal{L}$-1 norm of the pixel differences between 
the original image $\bm{x}$ and reconstructed image $\bm{\hat{x}}$, i.e., $\| \bm{x} - \bm{\hat{x}} \|_{1}$. Using the 
same number of hidden units, the errors for various training methods, which were averaged over
10,000 MNIST test samples and 3 independent experiments, are reported in Table. \ref{imputation}. 
As we can see, VPF performs much better than CD-1, CD-10 and PCD-1. It is also better than PCD-10 in most of the cases. Although PCD-10 is
slightly better than VPF in the Right case, it takes a longer time in doing 10 Gibbs transitions during training and 1000 during reconstruction.

\begin{table}[]
\centering
\caption{Average reconstruction error for 10,000 corrupted MNIST digits. 12 rows or columns out of 28 are randomly corrupted under 4 patterns, e.g., Top means top 12 rows corrupted. We tested 1 or 10 Gibbs updates for CD and PCD.}
\label{imputation}
\begin{tabular}{l||l|l|l|l||l}
\hline 
Corruption & CD-1 & CD-10 & PCD-1 & PCD-10 & VPF \\ \hline \hline
Top & 57.8 & 55.7 & 57.6 & 47.9 & \textbf{32.9} \\ 
Bottom & 74.7 & 54.5 & 58.2 & 48.9 & \textbf{46.9} \\ 
Left & 59.2 & 58.2 & 46.2 & 39.1 & \textbf{36.7} \\ 
Right & 69.9 & 55.4 & 53.6 & \textbf{43.7} & {44.1} \\ \hline
\end{tabular}
\end{table}

\begin{figure}[t]
  \centering
  \subfigure{\includegraphics[scale=0.5]{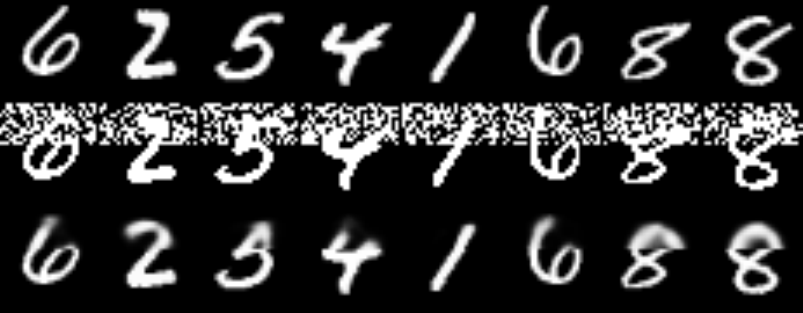}} 
      \subfigure{\includegraphics[scale=0.5]{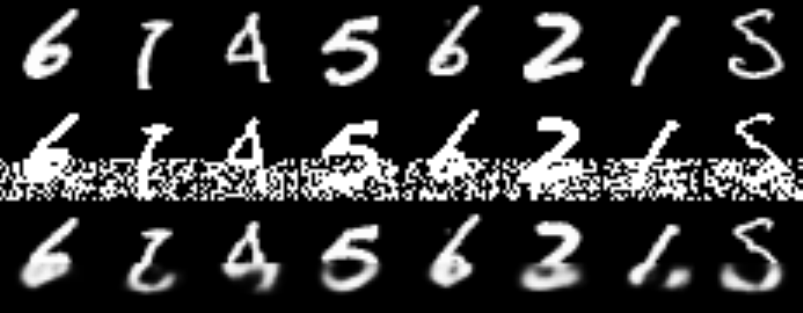}}
    \subfigure{\includegraphics[scale=0.5]{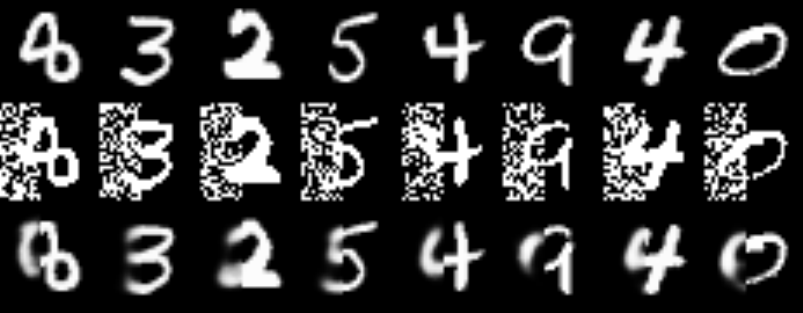}} 
     \subfigure{\includegraphics[scale=0.5]{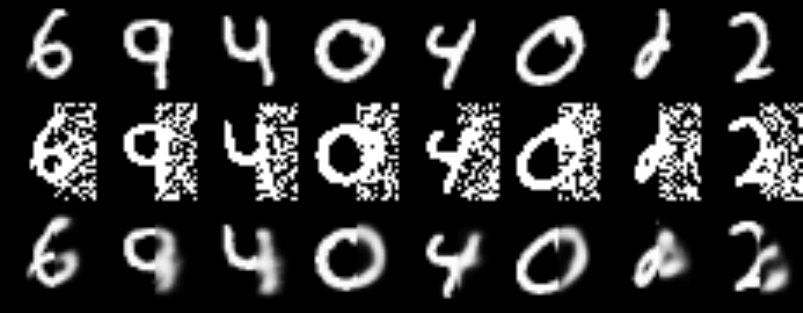}}         
   \caption{Examples of reconstructing the corrupted MNIST digits by RBM(196). 
   Top row: original digits, middle row: corrupted digits, bottom row: reconstructed digits.}
  \label{reconstruction}
\end{figure}

\subsection{Deep Boltzmann Machines}
\begin{figure*}[t]
  \centering
  \subfigure[DBM1 (left) \& DBM2 (right)]{\includegraphics[scale=0.35]{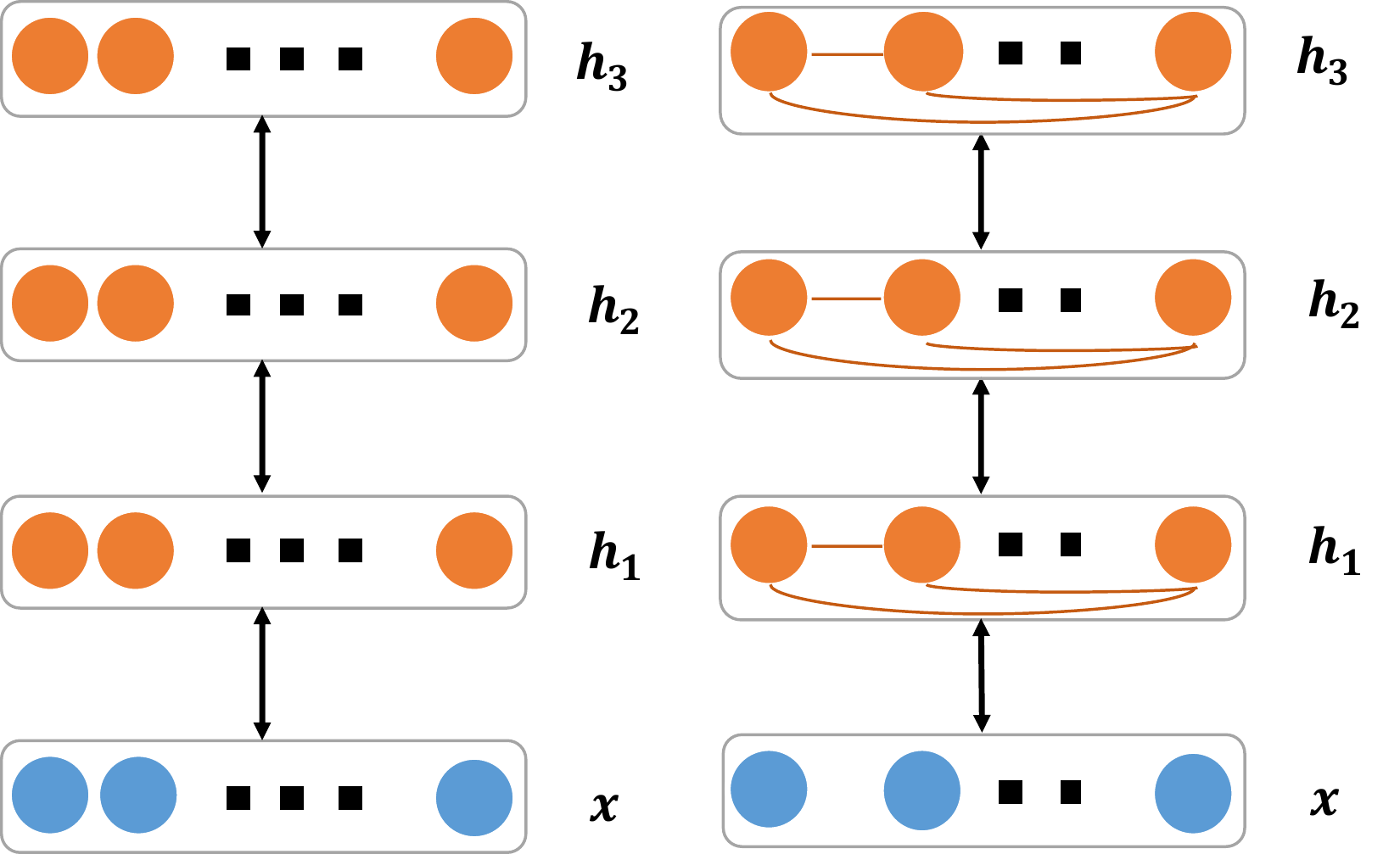}}  
  \subfigure[DBM1-P]{\includegraphics[scale=0.45]{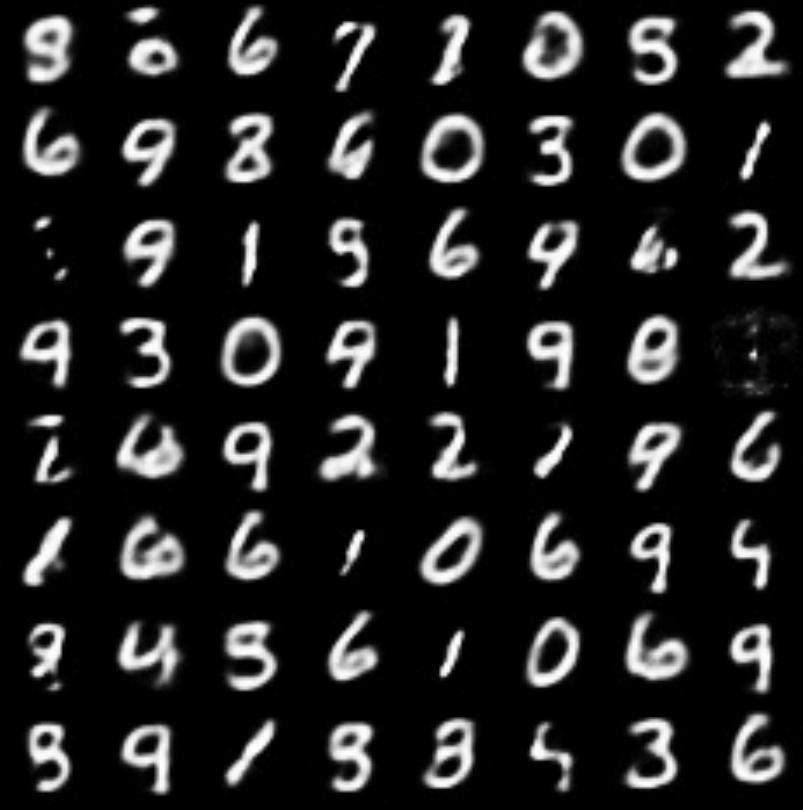}}   
  \subfigure[DBM2-R]{\includegraphics[scale=0.45]{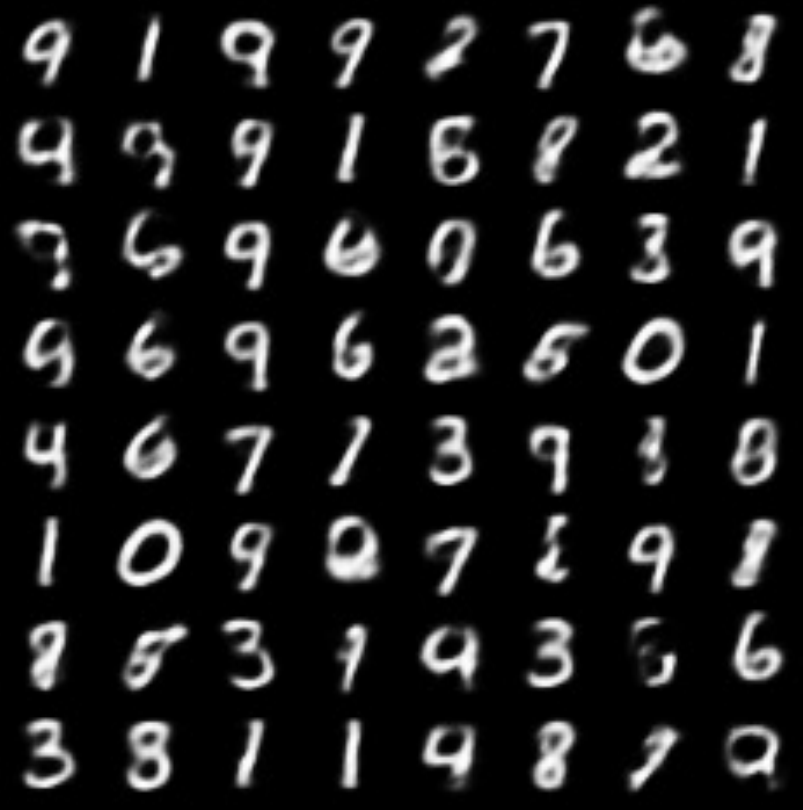}}   
  \subfigure[DBM2-P]{\includegraphics[scale=0.45]{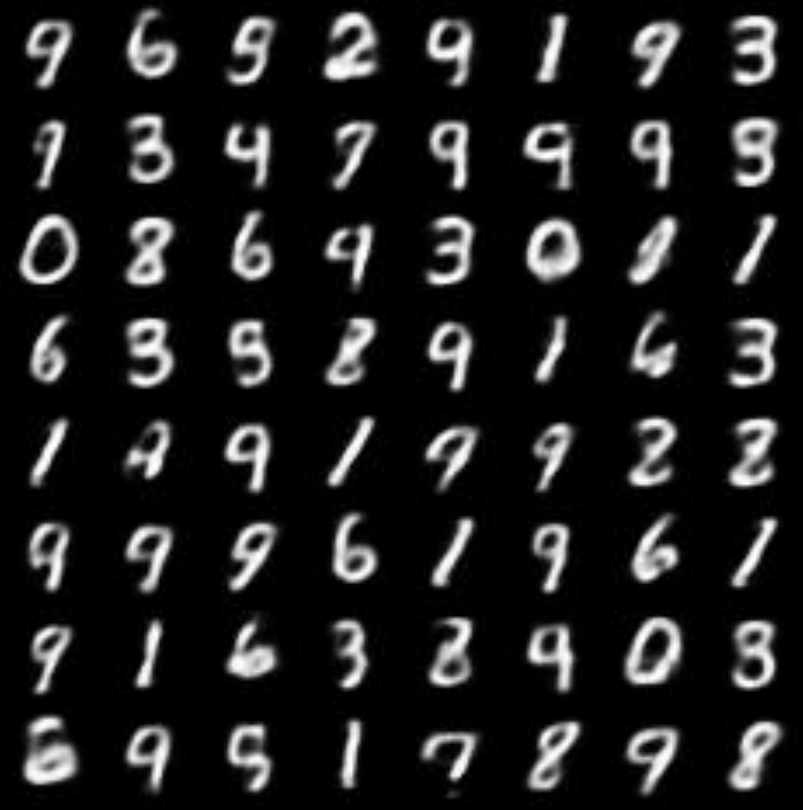}}      
   \caption{\textbf{(a)}. Different DBMs: left is traditional DBM, right is DBM with intra-layer connections. \textbf{(b-d)}. Digits generated from DBMs with different initialization for the top hidden layer, 
   not cherry-picked. E.g., DBM2-P refers to a DBM with intra-layer connections, generating samples 
with a mean activation prior. Greyscale images represent the probabilities for the visible units.}
  \label{generation}
\end{figure*}

The performance of VPF to train DBMs is evaluated on two datasets: MNIST and Fashion MNIST. For each dataset, 
we trained two DBMs with the same architecture 784-196-196-64 based on Algorithm. \ref{VPFDBM}. 
The first DBM (DBM1) has no intra-layer connections while the second one (DBM2) does. See Fig. \ref{generation}.(a) for an illustration. 

Algorithm. \ref{DBM-confab} is essentially a top-down approach for generating confabulations. To first initialize the top hidden layer $\bm{h}_\ell$, we experimented with two different strategies: (1) Randomly set states in $\bm{h}_\ell$ to be 0 or 1 
with equal probability (denoted as DBM-R), (2) Initialize $\bm{h}_\ell$ with a Bernoulli prior 
which is computed as the mean feedforward activations of the training dataset (denoted as DBM-P). We conjecture that in the absence of intra-layer connections, the bottom-up signals are crucial for initializing the factorial priors in a good neighborhood. This second strategy was used by \cite{bengio2015towards} as well for the generative task.

Confabulations generated from the trained DBMs using Algorithm. \ref{DBM-confab} are displayed in Fig. \ref{generation} and Fig. \ref{fashion}. 
We also estimated the log-likelihood of 10,000 MNIST test samples by fitting a Gaussian Parzen density
window to 10,000 samples generated by the learned DBMs \cite{goodfellow2014generative}. 
The standard deviation 
for the Parzen density estimator is chosen as 0.2 using a validation set
 following the procedure in \cite{desjardins2010parallel}. 
 The estimated log-likelihoods are reported in Table. \ref{lld}. 
 
 \begin{table*}[]
\small
\centering  
\caption{Log-likelihood and standard error estimated with Parzen density window. The reference measure obtained from MNIST training dataset is $239.88 \pm2.3 $ \cite{desjardins2010parallel}.     }
\label{lld}
\begin{tabular}{cccccc|lcccc}
\hline
DBM & DBN & Stacked-CAE & Deep GSN & GAN & \citeauthor{bengio2015towards} & \multicolumn{1}{c}{RBM(196)} & DBM1-R & DBM1-P & DMB2-R & DBM2-P \\ \hline
 32$\pm$2  & 138$\pm$2 & 121$\pm$1.6 & 214$\pm$1.1  &  225$\pm$2 & \textbf{236} & {78}$\pm$5.9 & {183}$\pm$1.1 & {210}$\pm$0.8& {222}$\pm$0.5  &222$\pm$0.4\\ \hline
\end{tabular}
\end{table*}

\subsubsection{MNIST.} From Table. \ref{lld} we clearly see that DBM1-P and DBM2-P are able to 
achieve log-likelihoods $LL= 210$ or $LL = 222$, which are much higher than RBM(196) $(LL=78$) because of the using of the high layers. They also perform better than many previous 
generative models such as deep Boltzmann machines  (DBM, $LL = 32$ )  \cite{bengio2014deep}, deep belief networks (DBN, $LL = 138$) \cite{bengio2013better}  and 
stacked contractive auto-encoders (Stacked-CAE, $LL = 121$) \cite{bengio2013better}.   While the log-likelihood of DBM1-P is slightly worse than that of deep generative stochastic 
nets (Deep-GSN, $LL =214$) \cite{bengio2014deep}, DBM2-P does better, thanks to its intra-layer connections. In fact, the log-likelihood of DBM2-P is quite close to that of generative adversarial nets 
(GAN, $LL = 225$) \cite{goodfellow2014generative},  and not far away from \cite{bengio2015towards} ($LL = 236$), which used a bottom-up signal to initialize the top layer.

As seen in Fig. \ref{generation}, trained DBMs generate realistic digit samples. 
We also note that DBM2 produced better 
samples than DBM1 (higher $LL$). While DBM1-R only achieved a log-likelihood of 183, 
DBM2-R achieved $LL = 222$ which outperforms DBM1-P and is as good as DBM2-P. This is consistent 
with our conjecture that intra-layer connections enable the learning of sparse distributed representations and highly-explanatory latent priors, so bottom-up signals are not required. Meanwhile, we found that DBM1 often
produced digits with undesirable gaps, see Fig.\ref{generation}.(b), which lower
the log-likelihood. We postulate that the gaps are due to DBMs exploring a factorial prior,
 a phenomenon commonly observed in GAN \cite{goodfellow2014generative}. This drawback 
 is resolved through the use of intra-layer connections which enable non-factorial priors. Indeed, there are no obvious gaps in digits from DBM2-R and DBM2-P.

\subsubsection{Fashion MNIST.} 
\begin{figure}[t]
  \centering
  \subfigure[Train samples]{\includegraphics[scale=0.32]{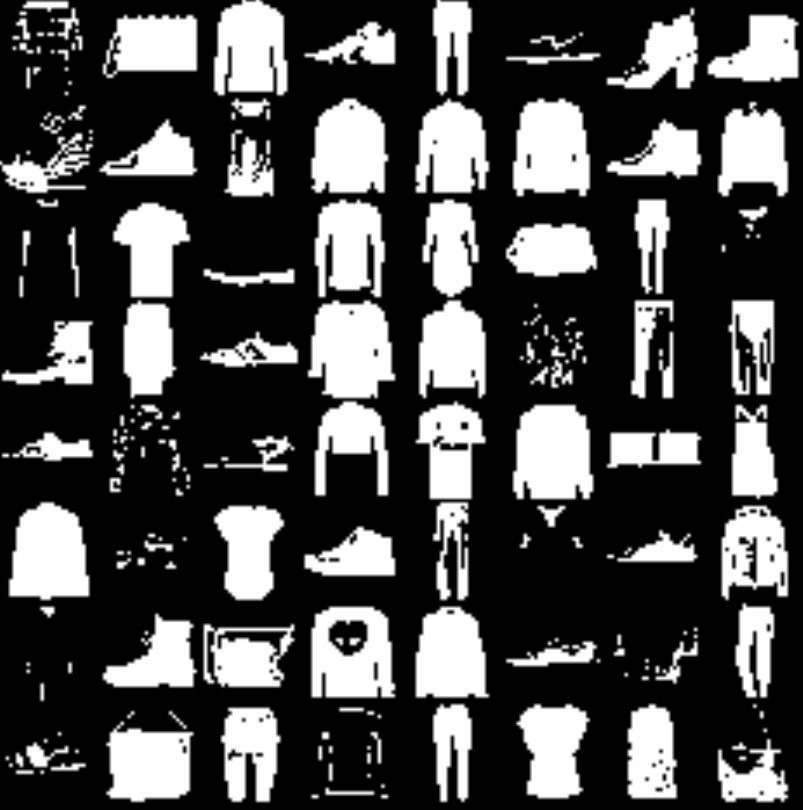}}   
  \subfigure[DBM1-P]{\includegraphics[scale=0.32]{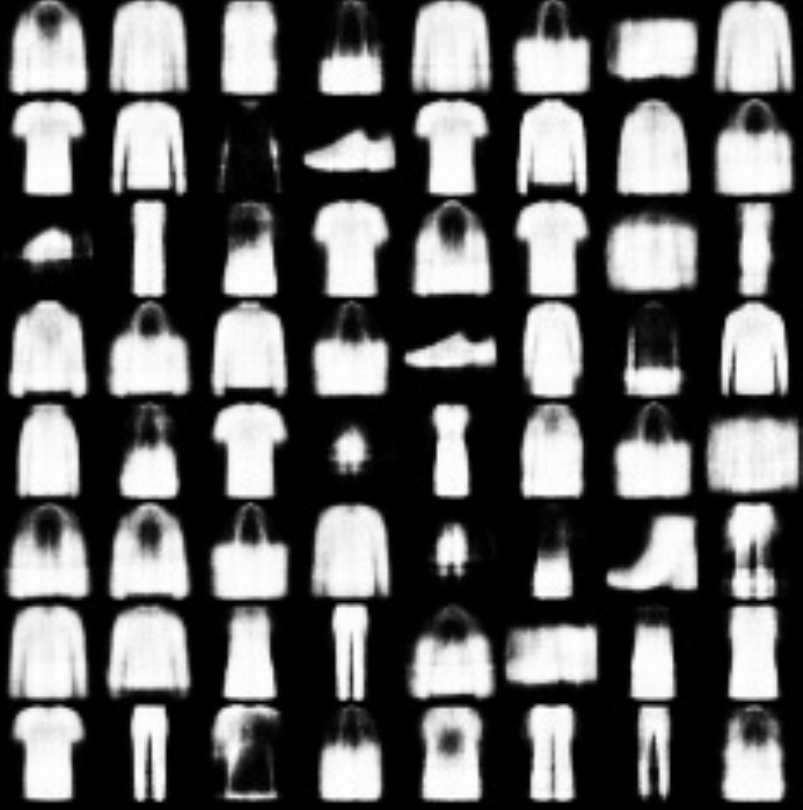}}        
  \subfigure[DBM2-P]{\includegraphics[scale=0.32]{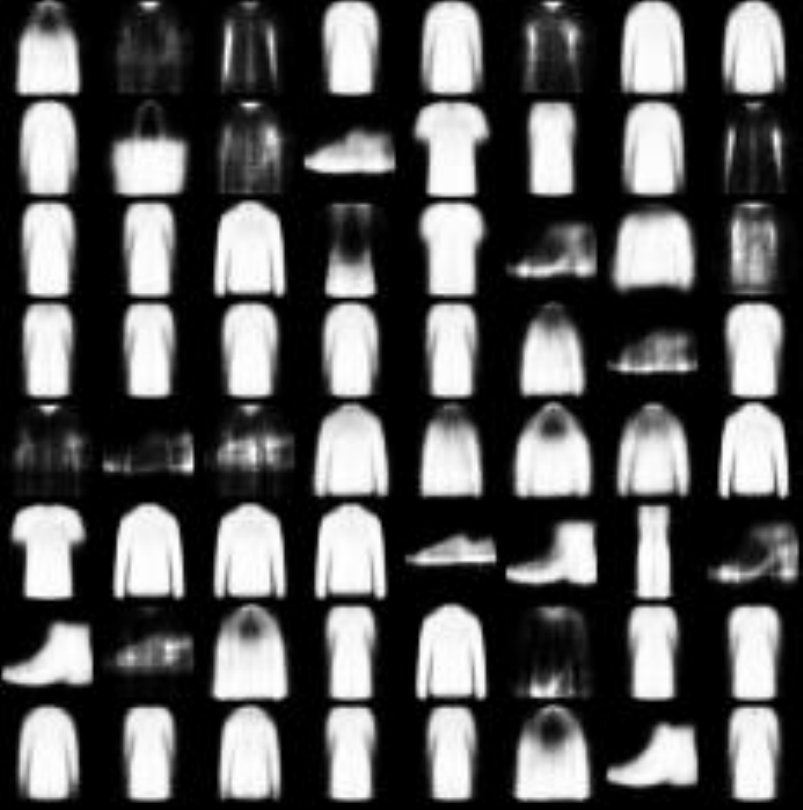}}   
   \caption{Generation of fashion images, not cherry-picked. Training samples are binarized by thresholding with $0.5$. }
  \label{fashion}
\end{figure}

We trained DBMs on Fashion MNIST, a data set that contains ten kinds of 
fashion images from Zalando \cite{xiao2017fashion}. 
As we can see from Fig. \ref{fashion}, the learned DBMs generate meaningful samples such as T-shirts, coats, and ankle boots, all without any other tricks for the learning procedure or hyper-parameters. 
The generated samples do not look as good as the samples for MNIST. One likely reason is that thresholding damages the Fashion MNIST data significantly. 
Samples from DBM1-R and DBM2-R are not shown here due to space constraints.

\section{Related Work \& Discussion}
Our work is highly related to that of \cite{bengio2015towards}. We both are able to train 
deep generative models without backpropagation, using variational ideas while making sense of STDP. 
However, our work differs from \cite{bengio2015towards} in two aspects. First, we use different weight update rules to explain STDP, as stated in the introduction. 
 Second, we employ a different strategy to avoid backpropagation, i.e., we use the MPF update rule instead of target propagation. 

Another similar work which also explores STDP using energy-based models is equilibrium propagation \cite{scellier2017equilibrium}. 
In equilibrium propagation, the objective function is the sum over the data of $E+\beta C$ where $E$ is the energy of the network, 
$C$ is the loss between a model prediction and an output target, and $\beta$ a trade-off parameter. 
VPF is a generative rather than discriminative model, so it does not aim to minimize $C$. 
Instead of minimizing the network energy $E$, it optimizes the sum of $\exp(E(\bm{y})/2-E(\bm{x})/2)$ over all one-hop neighbors $\bm{y}$ of a data point $\bm{x}$. 
This objective is a kind of exponential transition (kinetic) energy rather than a form of state (potential) energy.

We compare VPF with existing methods to train RBMs or DBMs.
The differences between VPF and CD are: (1) VPF has an explicit objective function, (2) VPF does not require 
Gibbs sampling to generate confabulations to compute the gradient, (3) VPF gives rise to STDP. As for training DBMs, 
instead of employing a two-stage process, i.e., pre-training
a stack of RBMs or a separate recognition model to initialize a DBM and then updating it \cite{salakhutdinov2009deep,salakhutdinov2010efficient}, 
VPF learns all the parameters simultaneously without greedy layer-wise initialization of the weights and does not require data confabulations. Moreover, VPF trains neural nets with intra-layer connections, 
which is hard to achieve in conventional methods. 

Recently, many groups have worked on embedding STDP into deep learning for biologically plausible training of neural nets, e.g.,   
\cite{rezende2011variational} found STDP-like behavior in recurrent spiking networks, \cite{bengio2017stdp} 
proposed a framework for training energy-based models without explicit backpropagation. 
VPF is yet another stab in this direction. Even with these advancements, there is still more to be done. For instance, a key question is whether it is possible or even necessary to do away with symmetric weights in energy-based models, so as to achieve biological plausibility. Some studies suggest that methods such as batch-normalization could help ameliorate this ``weight transport problem" \cite{liao2016important}. 

\section{Conclusion \& Future Work}
\label{conclusion}

In this paper, we derived a new training objective and learning algorithm known as Variational Probability Flow, 
which gives rise to biologically plausible training of deep neural networks. 
There are three main directions for future work. Firstly, we are interested in extending VPF to recurrent spiking 
networks which we believe are closer to biological networks. We hope that by doing so, we also uncover a directed version of VPF whose 
weight updates are given simply by $\Delta w_{ij} \propto - y_i \alpha_j \delta_j$ to eliminate the requirement for symmetric weights. 
Secondly, while neurons compute using binary values, real-world data is mostly represented by real values. Therefore, 
we hope to design natural interfaces between the two data types for more efficient learning.
Finally, we will explore how the local learning rules may be exploited to design model-parallel algorithms that run efficiently on GPUs. 
Such algorithms may enable us to train massive neural networks with tens of billions of neurons in a distributed energy-efficient manner.

\section{Acknowledgement}
Shaowei Lin is funded by SUTD grant SRES15111 and SUTD-ZJU grant ZJURP1600103. 
Zuozhu Liu is supported by SUTD President's Graduate Fellowship. 
We would like to thank NVIDIA for their computational support. 
We also thank Christopher Hillar, Sai
Ganesh, Binghao Ng, Dewen Soh, Thiparat Chotibut, Gary
Phua, Zhangsheng Lai for their helpful work and discussions.

\bibliography{ref}

\bibliographystyle{aaai}

\newpage
\onecolumn

\begin{center}
\huge\textbf{Supplementary Material}
\end{center}

\section{Fully-Observed Boltzmann Machines}
\label{updaterule}

Given a training data point $\bm{y} \in \mathcal{D}$, let $\bm{x}$ be the one-hop neighbor of $\bm{y}$ that differs in the $j$-th bit, i.e. $x_i=y_i$ for all $i \neq j$, and $x_j = 1-y_j$. Let $g_{xy} = 1$. Then,
\begin{align*}
\Gamma_{xy} &= g_{xy} \exp \Big( \frac{1}{2} \text{Energy}(\bm{y} ;\theta) - \frac{1}{2}\text{Energy}(\bm{x} ;\theta) \Big)\\
&= \exp \Big( -\frac{1}{2} (\textstyle \sum_{m\neq n} w_{mn}y_m y_n  + \sum_{m} b_m y_m )+  \quad  \frac{1}{2} (\sum_{m\neq n} w_{mn}x_m x_n  + \sum_{m} b_m x_m )  \Big) \\
&= \exp \Big( \frac{1}{2} (\textstyle \sum_{i\neq j} w_{ij}(x_ix_j- y_i y_j)  + b_j (x_j-y_j))  \Big)\\
&= \exp \Big( \frac{1}{2} (x_j-y_j) (\textstyle \sum_{i\neq j} w_{ij}y_i  + b_j )  \Big) = \delta_j
\end{align*}
where $\delta_j = \exp(\alpha_j z_j)$, $\alpha_j=1/2-y_j$ and $z_j =  \sum_{i\neq j} w_{ij}y_i  + b_j$.

Now, for all states $\bm{x},\bm{y}$ of the Boltzmann machine, let the connectivity $g_{\bm{xy}}=1$ if and only if $\bm{x},\bm{y}$ are one-hop neighbors. We denote the one-hop neighors pf $\bm{y}$ by $\mathcal{N}(\bm{y})$. Using this set of connectivities and assuming that $\mathcal{N}(\bm{y})$ does not intersect $\mathcal{D}$ for all $\bm{y} \in \mathcal{D}$, the MPF objective function becomes
\begin{align*}
\mathcal{K}(\theta) &= \frac{\varepsilon}{N} \sum_{\bm{y} \in\mathcal{D}} \sum_{\bm{x} \in \mathcal{N}(\bm{y})} \Gamma_{xy} = \frac{\varepsilon}{N}  \sum_{k=1}^N \mathcal{K}^{(k)}(\theta), \quad \mathcal{K}^{(k)}(\theta) = \sum_{j=1}^{|\mathcal{V}|} \delta_j^{(k)}
\end{align*}
where $\delta_j^{(k)} =  \delta_j(\bm{y}^{(k)})$. Consequently, for each data point $\bm{y}^{(k)}$, the gradients of $\mathcal{K}^{(k)}$ are
\begin{align*}
\frac{\partial \mathcal{K}^{(k)} }{ \partial b_i} =  \alpha_i \delta_i, \quad \frac{\partial \mathcal{K}^{(k)}} { \partial w_{ij}} =  y_j \alpha_i \delta_i  + y_i \alpha_j \delta_j.
\end{align*}

\section{VPF Objective Function}

Given any two joint distributions $p(\bm{h},\bm{x})$ and $q(\bm{h},\bm{x})$, we have
\begin{align*}
D_{\text{KL}}(q(\bm{h},\bm{x}) \| p(\bm{h},\bm{x})) &= \int\!\!\int q(\bm{h},\bm{x}) \log \frac{q(\bm{h},\bm{x})}{p(\bm{h},\bm{x})}\,d\bm{h}d\bm{x} \\
&= \int\!\!\int q(\bm{x})q(\bm{h}|\bm{x}) \log \frac{q(\bm{x})q(\bm{h}|\bm{x})}{p(\bm{x})p(\bm{h}|\bm{x})}\,d\bm{h}d\bm{x} \\
&= \int\!\!\int q(\bm{x})q(\bm{h}|\bm{x}) \log \frac{q(\bm{h}|\bm{x})}{p(\bm{h}|\bm{x})}\,d\bm{h}d\bm{x}\,\,  +  \int\!\!\int q(\bm{h}|\bm{x})q(\bm{x}) \log \frac{q(\bm{x})}{p(\bm{x})}\,d\bm{h}d\bm{x}\\[4pt]
&=\mathbb{E}_{\bm{x} \sim q(\bm{x})} [D_{\text{KL}}(q(\bm{h}|\bm{x}) \| p(\bm{h}|\bm{x}))] \,\,+ D_{\text{KL}}(q(\bm{x}) \| p(\bm{x})).
\end{align*}
Substituting  $p(\bm{h},\bm{x}) =  p^{(t)}(\bm{h},\bm{x};\theta)$ and $q(\bm{x}) = \tilde{p}^{(0)}(\bm{x})$ gives Eq. 3 in the paper.

\newpage
\thispagestyle{empty}
\begin{table}[]
\centering
\caption{Average reconstruction error for 10,000 corrupted MNIST digits. 12 rows or columns out of 28 are randomly corrupted under 4 patterns.}
\label{imputation}
\begin{tabular}{l||l|l|l|l||l|l}
\hline 
Corruption & CD-1 & CD-10 & PCD-1 & PCD-10 & RBM(196) & BM(196)   \\ \hline 
Top & 57.8 & 55.7 & 57.6 & 47.9 & \textbf{32.9}& 34.33  \\ 
Bottom & 74.7 & 54.5 & 58.2 & 48.9 & \textbf{46.9}& 47.65  \\ 
Left & 59.2 & 58.2 & 46.2 & 39.1 &  36.7 & \textbf{36.41} \\ 
Right & 69.9 & 55.4 & 53.6 & 43.7 & {44.1}& \textbf{42.31} \\ \hline
\end{tabular}
\end{table}

\begin{figure}[t]
  \centering
  \subfigure{\includegraphics[scale=0.6]{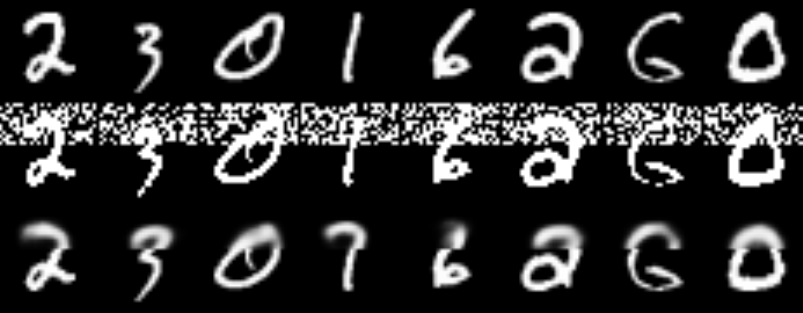}} 
      \subfigure{\includegraphics[scale=0.6]{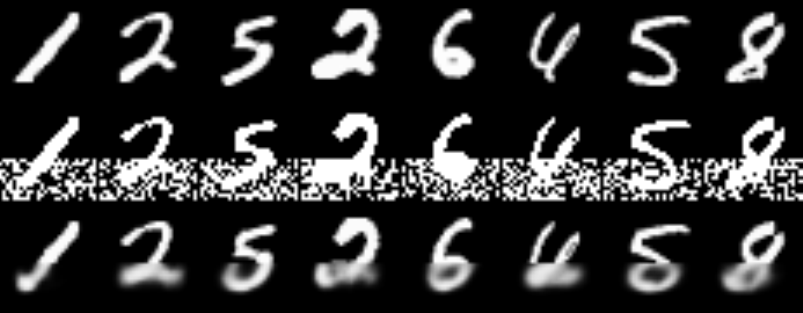}} \\
    \subfigure{\includegraphics[scale=0.6]{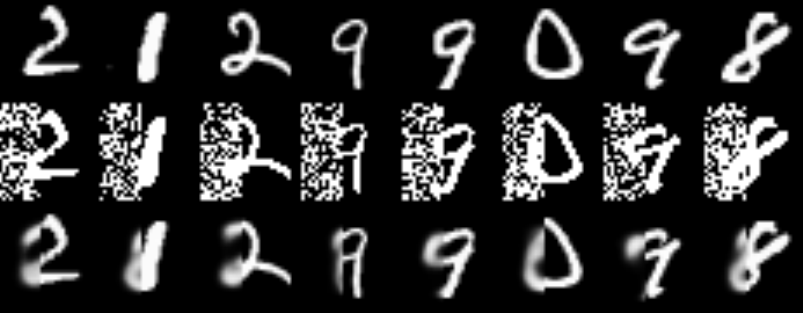}} 
     \subfigure{\includegraphics[scale=0.6]{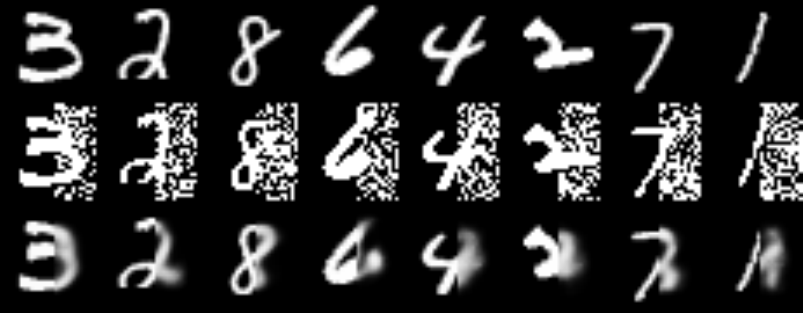}}         
   \caption{Examples of reconstructing the corrupted MNIST digits by BM(196). 
   Top row: original digits, middle row: corrupted digits, bottom row: reconstructed digits.}
  \label{reconstruction}
\end{figure}

\section{Experiment Details}

The code will be available at 
\begin{center}
 \url{https://github.com/owen94/Variational-Probability-Flow}. 
\end{center}


The hyper-parameters used in the experiments are specified in the main paper. Our implementation of the variational EM algorithm is described in the section "VPF for Multilayer Networks". During the E-step, Gibbs sampling is performed in each layer while zeroing out contributions from deeper layers. During the M-step, the parameters are updated according to Equations 1 and 2. Precautions are taken at every step to ensure that the weight matrix is symmetric and that the diagonal entries are zero.

Because our model includes intra-layer connections, we choose 196 hidden units to keep the number of parameters small. We also want to show that with VPF, small networks are able to generate good samples. As for the estimation of log-likelihood, we use the Parzen density estimator to be consistent with all the previous baselines, especially that of (Bengio et.al., 2015). As mentioned in (Goodfellow et.al., 2014), it is difficult to implement AIS for generative adversarial networks (GANs), so we did not use the AIS for comparison of our model with GANs. 

Due to space constraints, we provide in this appendix the results for image reconstruction for a Boltzmann machine trained via VPF with one hidden layer that has intra-layer connections. We use 196 hidden nodes to be consistent with the main paper. This model is denoted as BM(196). The results are shown in Table.~\ref{imputation} and Figure.~\ref{reconstruction}. Overall, the performance of BM(196) is very similar to that of RBM(196). In the ``Top" and ``Bottom" cases, RBM(196) performs slightly better, while in the ``Left" and ``Right" cases, BM(196) performs slightly better.

\end{document}